# Meta-path Analysis on Spatio-Temporal Graphs for Pedestrian Trajectory Prediction

Aamir Hasan, Pranav Sriram, and Katherine Driggs-Campbell

*Abstract*— Spatio-temporal graphs (ST-graphs) have been used to model time series tasks such as traffic forecasting, human motion modeling, and action recognition. The high-level structure and corresponding features from ST-graphs have led to improved performance over traditional architectures. However, current methods tend to be limited by simple features, despite the rich information provided by the full graph structure, which leads to inefficiencies and suboptimal performance in downstream tasks. We propose the use of features derived from meta-paths, walks across different types of edges, in ST-graphs to improve the performance of Structural Recurrent Neural Network. In this paper, we present the Meta-path Enhanced Structural Recurrent Neural Network (MESRNN), a generic framework that can be applied to any spatio-temporal task in a simple and scalable manner. We employ MESRNN for pedestrian trajectory prediction, utilizing these meta-path based features to capture the relationships between the trajectories of pedestrians at different points in time and space. We compare our MESRNN against state-of-the-art ST-graph methods on standard datasets to show the performance boost provided by meta-path information. The proposed model consistently outperforms the baselines in trajectory prediction over long time horizons by over 32%, and produces more socially compliant trajectories in dense crowds. For more information please refer to the project website at https://sites.google.com/illinois.edu/mesrnn/home

## I. INTRODUCTION

Pedestrians moving in crowds consider multiple factors such as the location of their intended goal, the obstacles that may be in their way, and the movement of other pedestrians in their vicinity. Recent works in pedestrian trajectory prediction have worked towards developing techniques that consider the above three factors [1]–[9]. However, these methods either consider other dynamic agents as obstacles by excluding their motion or use the data about their motion inefficiently. We aim to tackle this inefficiency by using features derived from meta-paths in a spatio-temporal representation of the problem.

Spatio-Temporal graphs (ST-graphs) have been widely used to improve the state-of-the-art in many fields within robotics such as surgical robotics, predictive planning, localization, and navigation [10]–[14]. Trajectory prediction is a cornerstone within these fields, driving tasks such as autonomous driving, robot navigation and decision systems [15]. These tasks require robots to have the ability to accurately predict the movement of other agents to plan their own movement [16]. Recently, ST-graphs based methods have been applied successfully in trajectory prediction [2], [3], [5], [8]. However,

The authors are with the Department of Electrical and Computer Engineering, University of Illinois at Urbana-Champaign, Champaign, IL 61820 {aamirh2, psriram2, krdc}@illinois.edu

This material is based upon work supported by the National Science Foundation under Grant No. 2143435.

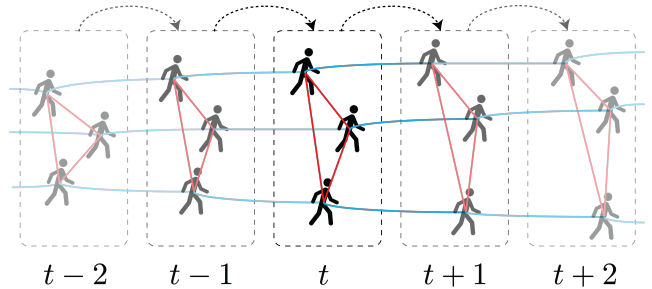

Fig. 1: A scene involving multiple pedestrians moving independently, yet aware of each other's movements.

these methods do not capitalize on all the information stored in the graphs and tend to only use a fraction of the features that can be derived from the structure. For example, ST-LSTM only uses the direct spatial information at a fixed time step in their model, ignoring the features that long term spatial dependencies in the representation provide [2].

Meta-path analysis has been widely leveraged in fields such as similarity searches and community detection [17], [18]. By explicitly capturing indirect and long-term relationships between nodes, meta-paths are able to provide rich features that inform downstream tasks [19].

We aim to exploit meta-paths to make efficient use of the ST-graph structure through a new model, which we call Meta-path Enhanced Structural Recurrent Neural Network (MESRNN). Using the features derived from different types of meta-paths as inputs, our model uses multiple RNN-based modules to encode and capture more information from the graph. In doing so, our model is able to gain a better understanding of the environment when provided with the same structural information as other techniques. We demonstrate our MESRNN framework on the trajectory prediction task, showing improved performance over state-of-the-art ST-graph techniques. The proposed model is generalizable, as MESRNN can be applied to any other spatio-temporal tasks, and scalable, as the number of features in the model is independent of the number of pedestrians in the scene.

Our main contributions are twofold: (1) We present a novel generic framework called Meta-path Enhanced Structural RNN, which utilizes meta-path based features to predict attributes in time series tasks that can be represented with ST-graphs; and (2) We apply meta-path analysis and MESRNN to the pedestrian trajectory prediction task, outdoing state-of-the-art ST-graph methods.

This paper is organized as follows. Section II reviews related work pertaining to meta-path analysis, ST-graphs,

and pedestrian trajectory prediction. Section III describes the problem statement, different constituent elements of MESRNN, and implementation details. Section IV describes the experiments on standard datasets for trajectory prediction. Section V discusses the results and compares the performance of our method with state-of-the-art baselines. Finally, we summarize our findings in Section VI.

## II. RELATED WORK

### A. Spatio-Temporal Graphs

Spatio-Temporal graphs have been not only been applied to aid with tasks in robotics, but are also extensively used in tasks outside robotics such as traffic forecasting, route planning, urban planning and public safety, and commerce [20]–[22]. A primary reason for their popularity is their general applicability in capturing and encoding the behavior of multiple interacting entities whose attributes change across space and time [22]. In a ST-graph, the vertices represent the attributes of the entities at different times in the scene. *Spatial* edges describe the spatial relationship between the vertices at the same time step. *Temporal* edges are defined to encapsulate the change in the vertices' attributes over consecutive time steps. Figure 2 shows the ST-graph representation of the scene with pedestrians that are shown in Figure 1.

### B. Meta-path Analysis

In a graph, a *walk* is a sequence of nodes where each successive pair of nodes is adjacent [23]. A *Meta-path* is a walk where successive edges need not be of the same type. The features derived from meta-path analysis are particularly useful in improving graph-based methods because they encode indirect semantic relations between two vertices that may not be directly connected to each other. The features also aid in establishing new relations between vertices. Thus, meta-path analysis has been used to to identify new gene set groups in gene networks [17], improve clustering algorithms [19], and in community detection in general heterogeneous graphs [18]. As such, meta-paths have been found to provide important insights in finding new patterns and relations in graphs.

### C. RNNs for Trajectory Prediction

Traditional methods for trajectory prediction consist of physics-based methods [24]–[27], which are over-constrained and over-complicated [28]. Learning-based methods, however, are data driven and can freely learn the subtleties in the space based on the observed data [29]. Methods that use Recurrent Neural Networks (RNNs) are of particular interest to the scientific community due to the popularity and utility of such models in other time series tasks [30].

RNN models have been used extensively in multiple areas such as acoustic modelling [31], speech recognition [32], [33], and natural language processing [34], [35]. Recently, RNNs have also been applied to trajectory prediction, in methods such as ST-LSTM [2], ST-Transformer [3], Social Attention [6], MIF-WLSTM [1] , and Social GAN [4]. However, most of these methods, *e.g.* SR-LSTM [7], do not use the information provided by the movements of all agents contained in the graph, and instead prefer for the networks to learn this information. Our goal is to leverage this information by explicitly using features based on the structured information provided by a spatio-temporal representation.

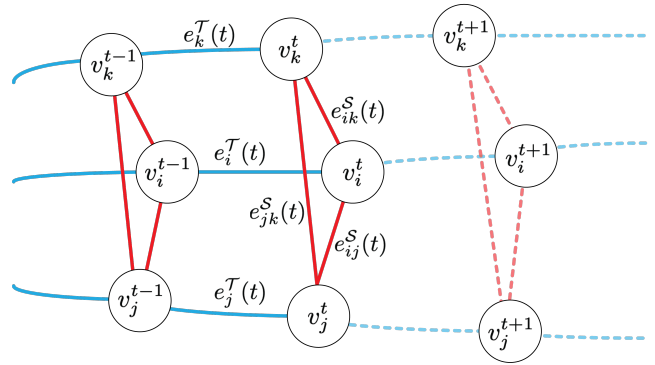

Fig. 2: The ST-Graph representation of a section of Figure 1. Red edges represent spatial edges and blue edges represent temporal edges. Edges represented as dashed lines are ones that are yet to be predicted while those represented by solid lines are edges which have already been observed or predicted.

## III. METHOD

In this section, we describe how a time series problem can be represented as a ST-graph. We also present the notion of meta-paths in ST-graphs and describe how we construct and train our MESRNN model. We showcase the above process in relation to the pedestrian trajectory prediction task, which can then be generalized to any other time series task.

We now formally define the pedestrian trajectory prediction task. Let $\{(x_i^t, y_i^t)\}$ represent the spatial coordinates of $N$ pedestrians, $i \in \{1, 2, \ldots, N\}$. Given the trajectories of the pedestrians observed for some period $T_{obs}$, $\{(x_i^t, y_i^t)\}_{t=1}^{T_{obs}}$, we want to predict the trajectories of the pedestrians for the next $T_{pred}$ time steps, $\{(x_i^t, y_i^t)\}_{t=T_{obs}+1}^{T_{obs}+T_{pred}}$.

### A. Representation of Spatio-Temporal Graphs

Based on the positions of the pedestrians, we construct a graph $\mathcal{G} = (\mathcal{V}, \mathcal{E}^S, \mathcal{E}^T)$. Here, $\mathcal{V}$ is the set of vertices, which represent the positions of pedestrians. $\mathcal{E}^S$ is the set of *spatial* edges connecting two vertices that represent the positions of *different* pedestrians at the *same time step*. $\mathcal{E}^T$ is the set of *temporal* edges connecting two vertices that represent the positions of the *same* pedestrian at *consecutive time steps*. We assume that all edges are undirected. We use the following notation to describe the graph elements:

$$\mathcal{V} = \{v_i^t\} \text{ where } v_i^t = (x_i^t, y_i^t)$$

$$\mathcal{E}^S = \{e^S{}_{ij}(t)\} \text{ where } e^S{}_{ij}(t) = d(v_i^t, v_j^t)$$

$$\mathcal{E}^T = \{e^T{}_i(t)\} \text{ where } e^T{}_i(t) = d(v_i^t, v_i^{t-1})$$

$$\text{for } t \in [1, T_{obs} + T_{pred}]; \ i, j \in [1, N]$$

where $d(\cdot, \cdot)$ is a distance function. Figure 2 shows the representation of the pedestrian trajectory prediction scene as a ST-graph with the notation as described above.

The Structural-RNN architecture further decomposes such an ST-graph into a factor graph [8]. The factor graph representation assists with training and provides a method to represent each pedestrian with a singular function by sharing the features of semantically similar factor vertices. That is, multiple spatial edge factor vertices, multiple temporal edge factor vertices, and multiple node factor vertices are each represented by a single RNN. Therefore, the number of learnable parameters in the model does not vary based on the number of nodes in the ST-graph or, in our case, the number of pedestrians in the scene, but only on the number of factors being considered. This condensation allows our method to be scalable irrespective of the number of pedestrians in the scene.

### B. Meta-paths in Spatio-Temporal Graphs

Our key insight into pedestrian (agent) interaction is that the future position (state) of an agent depends on the following factors: (1) how the position of other pedestrians has changed in the last few time steps; (2) how their position has changed from their previous position; and (3) how they are spatially related to the other pedestrians at the predicted time step. Our insight is motivated by the fact that humans observe other agents around them and make decisions with both spatial and temporal reasoning [28], (*i.e.*, people do not just focus on how far away other pedestrians are at the current time, but also take into account how others have been moving). Pooling modules have been used to incorporate this reasoning in previous works [4]–[6]. We incorporate this insight into our model via the concept of meta-paths in ST-graphs.

As described in Section II, a meta-path is a walk in a graph that can go across different types of edges. Formally, Consider a graph with vertices $\mathcal{V}$ and edges $\mathcal{E}$ with edge types $T$. A meta-path $\mathcal{P}$ of length $k$ is defined as a path between two vertices, $v_i$ and $v_j$, such that $v_i \xrightarrow{T_1} v_1 \xrightarrow{T_2} \ldots \xrightarrow{T_k} v_j$ ; where $v_1 \xrightarrow{T_1} v_2$ indicates that an edge of type $T_1$ connects vertices $v_1$ and $v_2$ for some $v_m \in \mathcal{V}$. The path $P$ is then said to be of type $T_1 - T_2 - \cdots - T_k$.

Generally, in any graph with $b$ edge types, for a given path length $l$, there are $b^l$ types of meta-paths. In ST-graphs, $b = 2$, as only spatial and temporal edges exist between vertices. Then, moving across a spatial edge followed by a temporal edge encodes information relating to how a pedestrian has moved across time with respect to their neighbor. This particular meta-path is called a Spatial-Temporal meta-path, one of the four meta-paths of length 2 in ST-graphs. The other three meta-paths are Spatial-Spatial, Temporal-Spatial, and Temporal-Temporal. These meta-paths can be observed in the ST-graph representation in Figure 2.

We can also construct meta-paths of longer lengths. But, in this paper, we restrict ourselves to meta-paths of length 2 as increasing the length of the meta-paths tends to increase the complexity of the model. Additionally, Sun *et al.* also found that meta-paths of shorter lengths tend to produce more useful features than longer meta-paths [36]. Note that meta-paths of length 1 are simply just the edges in the graph.

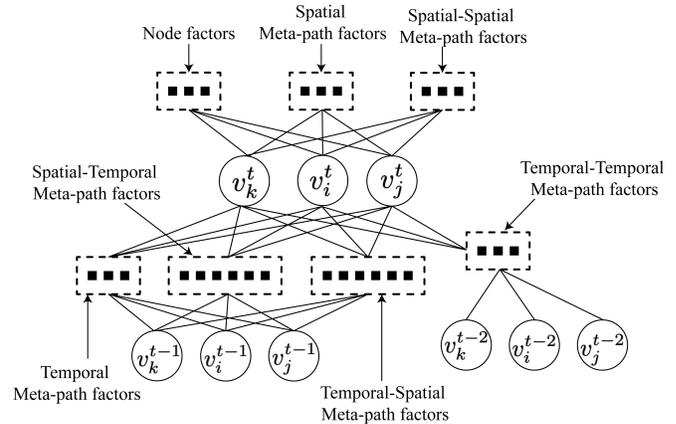

Fig. 3: A section of a simplistic representation of a factor graph constructed from an ST-graph with meta-path factors for meta-paths of length 1 and 2. Each square in the figure depicts a parameter associated with the nodes for each type of factor.

The encoding of each meta-path can be described in terms of their constituent edges as shown in Equations 1-4. Each type of meta-path provides a unique cognizance. Namely,

1) Spatial-Spatial meta-paths, $\mathcal{P}_{ij}^{\mathcal{SS}}(t)$, indicate how a pedestrian $i$ is located relative to pedestrian $j$, even if not connected explicitly via a spatial edge:

$$\mathcal{P}_{ij}^{\mathcal{SS}}(t) = e_{ik}^{\mathcal{S}}(t) \odot e_{kj}^{\mathcal{S}}(t) \text{ for some } k \neq i \neq j \quad (1)$$

2) Spatial-Temporal meta-paths, $\mathcal{P}_{ij}^{\mathcal{ST}}(t)$, indicate how a pedestrian $i$ is located relative to the past position of a pedestrian $j$:

$$\mathcal{P}_{ij}^{\mathcal{ST}}(t) = e_{ij}^{\mathcal{S}}(t) \odot e_{j}^{\mathcal{T}}(t) \quad (2)$$

3) Temporal-Spatial meta-paths, $\mathcal{P}_{ij}^{\mathcal{TS}}(t)$, indicate how a pedestrian $i$ was located relative to a pedestrian $j$ in the previous time step:

$$\mathcal{P}_{ij}^{\mathcal{TS}}(t) = e_{i}^{\mathcal{T}}(t) \odot e_{ij}^{\mathcal{S}}(t-1) \quad (3)$$

4) Temporal-Temporal meta-paths, $\mathcal{P}_{i}^{\mathcal{TT}}(t)$, indicate how pedestrian $i$ has moved across the last two time steps:

$$\mathcal{P}_{i}^{\mathcal{TT}}(t) = e_{i}^{\mathcal{T}}(t) \odot e_{i}^{\mathcal{T}}(t-1) \quad (4)$$

With this notion in mind, we now construct a new factor graph that includes these four new factors as shown in Figure 3. This factor graph representation becomes the basis for our MESRNN framework which is discussed in Section III-C.

### C. Meta-path Enhanced Structural RNN

To construct the MESRNN model, we replace each group of factor vertices with an RNN based submodule depending on the type of the factor. Figure 6 shows a broad overview of the MESRNN model. Each type of factor vertex is represented as a RNN. All the meta-path based factors are represented by EdgeRNNs (Figure 4) and the node factor is represented with the NodeRNN (Figure 5).

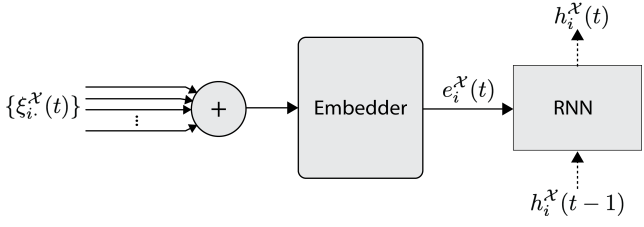

Fig. 4: An illustration of the EdgeRNN structure.

*a) EdgeRNN:* The EdgeRNN module takes all meta-paths of the same type that are connected to a vertex as input and outputs an embedding accounting for all meta-paths of that type. The EdgeRNN first sums up all the input meta-paths to produce one vector and then performs a simple embedding of that vector, which is then passed through a RNN. The RNN uses the embedding representing all the meta-paths of a type and the hidden state output from the previous time step to output the new hidden state. This hidden state representation is passed out as the output of the EdgeRNN. Equations 5 give a formal definition of the submodule. The module curates the number of inputs representing the neighbouring pedestrians into one single representation based on the type of meta-path that connects them. This submodule enables us to have a fixed number of inputs to the NodeRNN and also ensures that the model is scalable and does not have a variable number of features as the number of pedestrians in a scene varies. The learnable parameters in this submodule are the parameters used in the input encoder: $W^{\mathcal{X}}_{\text{encoder}}$, and the parameters in the RNNs: $W^{\mathcal{X}}_{\text{RNN}}$.

$$\xi^{\mathcal{X}}_{ij}(t) = \begin{cases} e^{\mathcal{X}}_{ij}(t), & \text{if } \mathcal{X} \text{ is a meta-path of length 1} \\ \mathcal{P}^{\mathcal{X}}_{ij}(t), & \text{if } \mathcal{X} \text{ is a meta-path of length 2} \end{cases} \quad (5a)$$

$$\begin{aligned}\xi^{\mathcal{X}}_i(t) &= \Sigma_j \xi^{\mathcal{X}}_{ij}(t) \\ e^{\mathcal{X}}_i(t) &= \phi\left(\xi^{\mathcal{X}}_i(t); W^{\mathcal{X}}_{\text{encoder}}\right) \\ h^{\mathcal{X}}_i(t) &= \text{RNN}\left(e^{\mathcal{X}}_i(t), h^{\mathcal{X}}_i(t-1); W^{\mathcal{X}}_{\text{RNN}}\right)\end{aligned} \quad (5b)$$

*b) NodeRNN:* The output embeddings from the EdgeRNN of each meta-path are taken as inputs to to the NodeRNN. The NodeRNN architecture first concatenates all the embeddings as well as an embedding for the position of the vertex. This concatenated embedding vector is then passed into an RNN. The RNN takes the hidden state output from the prediction at the previous time step along with the embedding vector as input to output the new hidden state. This hidden state representation is then decoded and output as an offset which is used to predict the position of the pedestrian represented by the vertex at that time step. Equations 6 give a formal definition of this submodule. This submodule is responsible for condensing all the information gathered from the ST-graph and predicting the output position. The learnable parameters in this submodule are the parameters used in the input embedder: $W^{\text{node}}_{\text{encoder}}$, the parameters in the RNN: $W^{\text{node}}_{\text{RNN}}$, and the parameters used for the offset decoder: $W^{\text{node}}_{\text{decoder}}$.

$$\begin{aligned}e_i(t) &= \phi\left(v^t_i; W^{\text{node}}_{\text{encoder}}\right) \\ m^1_i(t) &= [h^{\mathcal{S}}_i(t); h^{\mathcal{T}}_i(t)] \\ m^2_i(t) &= [h^{\mathcal{SS}}_i(t); h^{\mathcal{ST}}_i(t); h^{\mathcal{TS}}_i(t); h^{\mathcal{TT}}_i(t)]\end{aligned} \quad (6a)$$

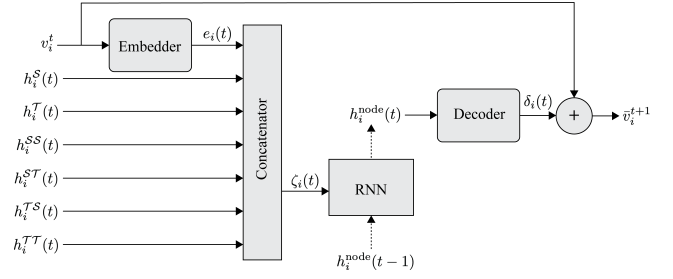

Fig. 5: An illustration of the NodeRNN structure.

$$\begin{aligned}\zeta_i(t) &= [e_i(t); m^1_i(t); m^2_t(t)] \\ h^{\text{node}}_i(t) &= \text{RNN}\left(\zeta_i(t), h^{\text{node}}_i(t-1); W^{\text{node}}_{\text{RNN}}\right) \\ \delta_i(t) &= \phi\left(h^{\text{node}}_i(t); W^{\text{node}}_{\text{decoder}}\right) \\ \bar{v}^{t+1}_i &= v^t_i + \delta_i(t)\end{aligned} \quad (6b)$$

*c) Training:* Figure 6 illustrates the entire MESRNN module using the EdgeRNN and NodeRNN submodules. The embedding layers in all submodules consist of a simple linear layer, followed by the tanh activation function and a dropout layer. The decoder in the NodeRNN is similar to the embedders except that it does not contain a dropout layer. All pedestrians share the features in all submodules.

For each time step, the module takes in the vertex's current position as well as the meta-paths connected to the vertex as inputs to the EdgeRNN submodules and the NodeRNN submodule. The model then predicts the position of the pedestrian represented by the current vertex and outputs the hidden states for all the recurrent components in the module to be used at the next time step.

For the duration of the observed period, the model uses the inputs from the ST-graph to obtain the current positions of the pedestrians in order to initialize the hidden states and learn their current movement. The module predicts the position for all pedestrians at a time step before moving on to predicting the position of any pedestrian at the next time step since the meta-path representations change based on current believed positions for all pedestrians. An alternative method to this, such as the one used by Vemula *et al*. is to use the data from the last observed period as input about the relationship between the different pedestrians for the whole predicted period [5]. While the latter would be less computationally intensive and faster, it would effectively ignore the subtleties in motion that the module would learn and output in its predicted positions about how the pedestrian's motion changes based on their neighbours' motion.

## IV. EXPERIMENTS

In this section we discuss the datasets we used to measure the performance of MESRNN and compare it against state-of-the-art methods. We also discuss the feature dimensions and hyper parameters used in our experiments, as well as the different metrics that were used to evaluate the trajectories predicted by the model.

### A. ETH/UCY Datasets

We evaluate our model on the publicly available ETH [37] and UCY [38] datasets. The datasets consists of five subsets:

ETH-UNIV, ETH-HOTEL, UCY-ZARA01, UCY-ZARA02, and UCY-UNIV. The trajectories are sampled at a rate of 2.5 frames per second. We preprocessed the raw datasets[1] and procured 3837 8-second long scenes with multiple pedestrians exhibiting various types of interactions and movement. Like other methods, we also used the leave one out testing strategy *i.e.*, the model is trained using scenes from four subsets and tested on scenes from the left out subset. We predicted the trajectories for all pedestrians present during the whole observed period of each scene. This approach was used for all baselines.

### B. Implementation Details

The trajectories were observed for 3.2-seconds and predicted for the next 4.8-seconds. For the prediction period, during training, we use Teacher Forcing by using the input positions from the ST-graph to predict the next positions. During inference, we use the position output in the previous time step by the model as input to obtain the current position of the pedestrian.

We used a 80-20 training-validation split on the training dataset. All trajectories were normalized to a range of [-1, 1] using min-max normalization in each dimension before any computation took place and were un-normalized after prediction.

All models were trained for 10 epochs and confirmed to have converged via their training loss characteristic. Training was performed with the ADAM optimizer [39] with a learning rate of 0.001. We used the simple difference operator as the distance function, $d$, to calculate the meta-path features. We used the Mean Square Error (MSE) loss function during training to help the model learn how to predict the output positions correctly. The global norm of the gradients was clipped at 10 after each training cycle. We did not use batch training as different training samples had a different number of pedestrians, which required that each scene be processed separately. All experiments were carried out on the HAL Cluster [40] using a single Nvidia V100 GPU.

We used Long Short-Term Memory (LSTM) [41] as the RNN unit. The hidden and cell states of the NodeRNN had a dimension of 256, while those of the EdgeRNNs had a dimension of 128. The embedder in the NodeRNN provides an embedding of length 128, where as the embedders in the EdgeRNNs provide embeddings of length 64.

### C. Baselines

We compare our model against two baseline models: (1) *Vanilla LSTM (VLSTM)* and (2) *Structural-RNN (SRNN)*. *Vanilla LSTM* is a simple LSTM network that takes in the current position and predicts the output position. We chose this model as our basic baseline to compare with over simpler models such as a Linear model and older models such as Social Force as earlier works have demonstrated that a simple LSTM model performs better or on par with them [6]. *Structural-RNN* is the model proposed by Jain *et al.* [8]. This model is similar to ours except that it does not use any information from the meta-paths.

### D. Evaluation Metrics

The results from the experiments were evaluated on the following standard metrics:

1) *Average Displacement Error (ADE)*: Average of the L2 distance between the predicted and ground truth trajectories at each time step of the prediction period. A lower value indicates that the predicted trajectory has a lower drift from the ground truth trajectories. Note that the observed positions of the pedestrians are not included while calculating this metric.
2) *Final Displacement Error (FDE)*: The L2 distance between the predicted and ground truth trajectories at the final time step of the prediction period. A lower value indicates that the predicted endpoint was close to the ground truth endpoint.

## V. RESULTS

In this section we analyze the trained model and baselines' performances on the ETH/UCY dataset and discuss their predicted trajectories.

### A. Quantitative Analysis

Table I provides a quantitative overview of our experiments. The simple Vanilla LSTM model performs the worst out of all the models tested. The model underperforms in the scenes from the UCY dataset when compared to the scenes from the ETH dataset. As noted by the authors of Social LSTM, this is likely due to the increased crowd density in the UCY dataset compared to the ETH dataset [6]. SRNN performs significantly better than VLSTM. However, the model still has the same issue as VLSTM, where the trajectories predicted in more crowded scenes, such as the ones from the UCY dataset, have higher error.

[1] https://github.com/erichhhho/DataExtraction

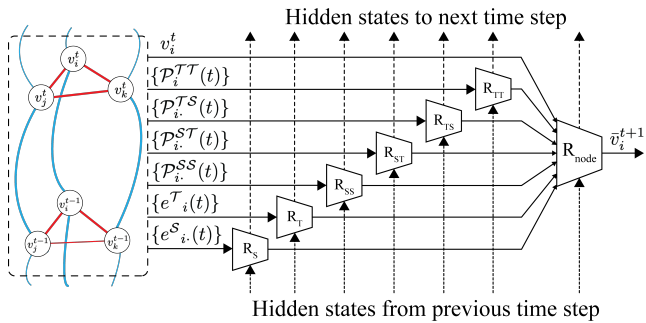

Fig. 6: The MESRNN model. The figure also shows the flow of data while MESRNN calculates the positions for pedestrian $v_i$ in the graph.

TABLE I: Results on the ETH/UCY Datasets

| Dataset | VLSTM | | SRNN | | MESRNN | |
|---|---|---|---|---|---|---|
| | ADE | FDE | ADE | FDE | ADE | FDE |
| ETH - UNIV | 0.0305 | 0.0721 | 0.0153 | 0.0326 | **0.0126** | **0.0257** |
| ETH - HOTEL | 0.0370 | 0.0912 | 0.0125 | 0.0278 | **0.0106** | **0.0215** |
| UCY - ZARA01 | 0.0594 | 0.1572 | 0.0220 | 0.0557 | **0.0076** | **0.0164** |
| UCY - ZARA02 | 0.0404 | 0.1037 | 0.0173 | 0.0427 | **0.0114** | **0.0272** |
| UCY - UNIV | 0.0485 | 0.1191 | 0.0094 | 0.0212 | **0.0093** | **0.0210** |
| Average | 0.0432 | 0.1087 | 0.0153 | 0.0360 | **0.0103** | **0.0224** |

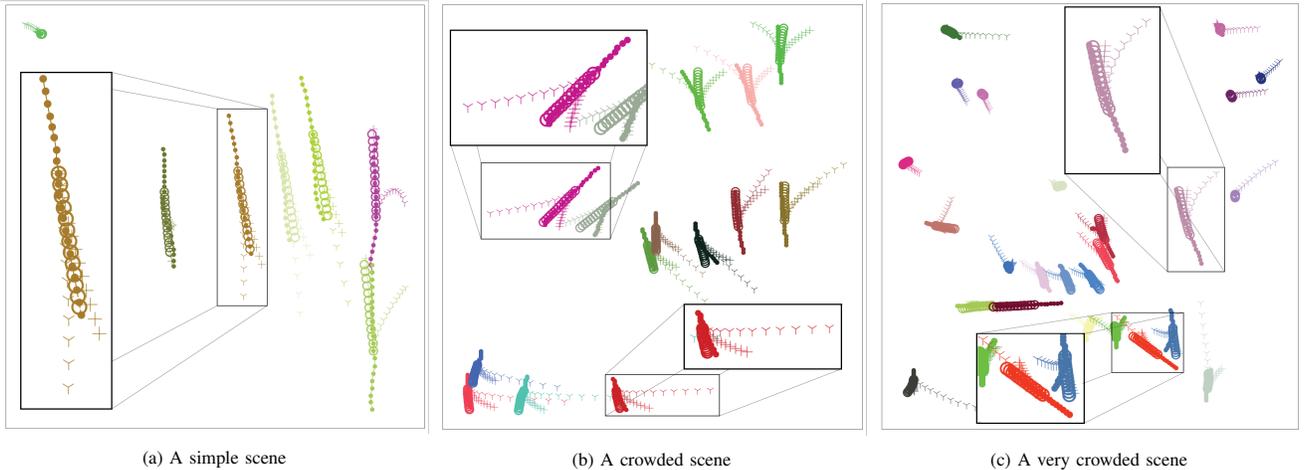

(a) A simple scene    (b) A crowded scene    (c) A very crowded scene

Fig. 7: Three scenes sampled from the dataset showing the trajectories predicted by the models. Each pedestrian is assigned their own color and the scene is shown to scale. The continuous dotted lines represent the observed period, the dashed dotted lines represent the ground truth trajectories, the lines drawn with the '∘' symbol represent the trajectories predicted by MESRNN, the lines drawn with the '+' symbol represent the trajectories predicted by SRNN, and the lines drawn with the '⊤' symbol represent the trajectories predicted by VLSTM. Trajectories of particular interest are overlaid on the scene.

Our MESRNN model significantly outperforms both baseline models, achieving a 75% and 32% improvement in ADE over VLSTM and SRNN respectively. Unlike VLSTM and SRNN, our model performs better in the densely crowded scenes from the UCY dataset than in the scenes from the ETH dataset. This improvement can be attributed to the inclusion of the numerous meta-path features due to the presence of a larger crowd in the scenes from the UCY dataset. However, this is not to say that the model does not perform well in the absence of data from neighbouring pedestrians, as the model still outperforms both baselines on the scenes from the ETH dataset. The consistent improved performance on both datasets validates our hypothesis that features derived from meta-paths significantly boost performance.

### B. Qualitative Analysis

Figures 7a, 7b, and 7c show the ground truth trajectories and the trajectories predicted by all the models. As seen in the figures, MESRNN tends to retain information from the observed trajectory such as the intended direction of motion and pace of movement better than the baseline models. This retention is responsible for the model predicting the final positions of the pedestrians more accurately, which is corroborated by the low FDEs observed in Table I. MESRNN is also able to predict the trajectories of agents who do not move significantly during the period much better than the baselines as observed in Figure 7c. Additionally, the model is able to adapt by changing the direction of motion and pace when more pedestrians are present in the surrounding area, something that the baseline models lack. This adaptability, that can be clearly observed in Figure 7b, is caused by the inclusion of meta-path based features which enable the model to predict better, more socially compliant trajectories in crowded spaces.

Overall, MESRNN exhibits increased performance over the baseline models without additional training samples or epochs and by only leveraging the additional features from meta-paths. Although MESRNN has increased computational costs due to the overhead of calculating the meta-path features, the training time is still achievable on a single GPU and remains efficient at runtime. The training time can be considerably reduced by using stronger compute resources. Hence, we can conclude that the inclusion of meta-path based features provides a significant boost in what models can learn from ST-graph representations.

## VI. CONCLUSION

In this paper, we present MESRNN, a novel framework that uses meta-path based features in ST-graphs to enhance the Structural RNN framework. We demonstrate the power of our method on the trajectory prediction task, outperforming baseline methods on a popular benchmark. Our use of meta-path based features provided a significant improvement over current methods that utilize ST-graphs.

While this initial work has demonstrated how additional graphical information (represented through meta-paths) is useful for enhancing performance, there are a number of directions to improve and explore. First, longer meta-paths and their impact on computation should be explored. In initial studies, the added computational complexity presented a challenge, but could be mitigated through intelligently selecting different types of meta-paths based on their added semantic value. Second, currently MESRNN only predicts one possible output trajectory for every input trajectory. While these trajectories are mostly accurate, they do not portray an accurate representation of the variable nature of human movement. Therefore, adapting the model to be able to predict multiple possible output trajectories, perhaps through predicting a probability distribution would be desirable. Additionally, our model should be tested in a real world experiment to aid with robot path planning. Lastly, the performance can be improved through use of goal information, adding additional structure to the prediction.